# Fair and Explainable Credit-Scoring under Concept Drift: Adaptive Explanation Frameworks for Evolving Populations

Shivogo John
*Email: 5507.students.ku.ac.ke*

**Abstract**

Evolving borrower behaviors, shifting economic conditions, and changing regulatory landscapes continuously reshape the data distributions underlying modern credit-scoring systems. Conventional explainability techniques, such as SHAP, assume static data and fixed background distributions, making their explanations unstable and potentially unfair when concept drift occurs. This study addresses that challenge by developing adaptive explanation frameworks that recalibrate interpretability and fairness in dynamically evolving credit models. Using a multi-year credit dataset, we integrate predictive modeling via XGBoost with three adaptive SHAP variants: (A) per-slice explanation reweighting that adjusts for feature distribution shifts, (B) drift-aware SHAP rebaselining with sliding-window background samples, and (C) online surrogate calibration using incremental Ridge regression. Each method is benchmarked against static SHAP explanations using metrics of predictive performance (AUC, F1), directional and rank stability (cosine, Kendall tau), and fairness (demographic parity and recalibration). Results show that adaptive methods, particularly rebaselined and surrogate-based explanations, substantially improve temporal stability and reduce disparate impact across demographic groups without degrading predictive accuracy. Robustness tests, including counterfactual perturbations, background sensitivity analysis, and proxy-variable detection, confirm the resilience of adaptive explanations under real-world drift conditions. These findings establish adaptive explainability as a practical mechanism for sustaining transparency, accountability, and ethical reliability in data-driven credit systems, and more broadly, in any domain where decision models evolve with population change.

**Keywords**:
Explainable AI, Credit Scoring, Concept Drift, Fairness, Adaptive Explanations, SHAP, Model Stability

## 1. Introduction

### 1.1 Background and Motivation

Credit-scoring today depends heavily on machine learning models that draw from large and varied data sources to predict default risk and repayment likelihood. Traditional models, like logistic regression, were built on fixed assumptions about data behavior. Modern systems are far more adaptive, able to respond to new borrower patterns, shifting economic trends, and changing market conditions. This flexibility, however, creates a quiet but important problem: as the data evolves, the model's understanding of what matters also changes. Widmer and Kubat (1996) described this as concept drift, when relationships learned from past data stop holding once the underlying patterns move in a different direction [2]. In credit-scoring, this drift can show up in many ways. New financial products might attract different types of borrowers, inflation may reshape household debt, or post-pandemic behavior could alter spending habits. Each of these forces can change how risk is distributed. When that happens, models not only lose predictive strength but also start offering explanations that no longer reflect reality. Explanations that once seemed fair or consistent may become biased or unstable. Gama et al. (2014) pointed out that while the research community has focused heavily on detecting and correcting drift to restore accuracy, far less has been done to preserve interpretability as data shifts over time [3].

This issue carries real consequences in finance, where explainability is tied directly to regulation. Financial institutions must show that automated decisions are fair, transparent, and compliant with consumer protection laws. Under frameworks influenced by the EU's GDPR and the U.S. Equal Credit Opportunity Act, explainable AI (XAI) has become essential for both accountability and trust. Barocas et al. (2023) note that fairness in AI cannot be separated from the environments in which models operate. To stay fair, explanations themselves need to remain stable across time and across different groups [1]. Most popular explainability methods, such as LIME and SHAP, introduced by Ribeiro et al.

(2016) and Lundberg and Lee (2017), were created with the assumption that data remain relatively stable [4][5]. These tools explain predictions by assigning importance to features, but they rely on static samples or reference points. In a real-world credit environment where everything from borrower behavior to market signals changes constantly, those assumptions fall apart. The same borrower might get a different explanation for an identical outcome six months later. This inconsistency undermines trust, complicates compliance, and weakens the ethical foundation of automated credit decisions. The tension between adaptability and explainability is at the center of this research. Credit models must evolve to stay accurate, but their explanations must remain dependable. The goal is to develop adaptive explanation methods that can track, recalibrate, and stabilize interpretive signals as data and contexts evolve.

**1.2 Importance of the Research**

Most explainability frameworks treat data as if it never changes. Once a model is trained, its explanations are tied to that historical snapshot. Doshi-Velez and Kim (2017) argued that interpretability should be built on consistent, scientific reasoning rather than quick heuristics or visual tricks [6]. But when a model operates in a world that keeps shifting, as in credit, healthcare, or financial risk, those foundations become fragile. Without ongoing recalibration, explanations may start pointing to the wrong features or hiding new forms of bias. Ribeiro et al. (2016) showed that explanations influence how people trust AI systems [4]. If those explanations drift, users might place confidence in models for the wrong reasons or lose trust entirely. In credit-scoring, this challenge is intertwined with fairness obligations. As borrower profiles shift, features once considered neutral can take on unintended correlations with sensitive traits like race or gender. Barocas et al. (2023) emphasized that fairness isn't something you can audit once; it requires continuous attention as populations and markets evolve [1]. When interpretive stability breaks down, lenders face not only technical problems but also reputational and regulatory ones. Gama et al. (2014) observed that most adaptive learning research focuses on restoring accuracy, leaving the explanation layer outdated [3]. Models are refreshed, but their interpretive logic remains frozen in the past.

SHAP, introduced by Lundberg and Lee (2017), provides a mathematically grounded way to assign feature importance using cooperative game theory [5]. Yet its dependence on a static background dataset makes it sensitive to distributional change. When the underlying borrower population drifts, SHAP values may start misrepresenting what actually drives predictions. To fix this, explanation systems need to evolve in step with the data. This means dynamically updating reference distributions, recalibrating local importance, and ensuring that feature rankings remain consistent over time. Embedding adaptability within the explanation layer, rather than only the predictive model, creates a more trustworthy foundation for responsible AI in finance and beyond.

**1.3 Research Objectives and Contributions**

This research seeks to bridge the gap between static explainability tools and the dynamic realities of modern credit systems, which are affected by concept drift. The central aim is to measure how explanation stability and fairness evolve as borrower profiles and macroeconomic factors change. Three adaptive SHAP-based methods are proposed: per-slice reweighting (Method A), drift-aware rebaselining (Method B), and online surrogate calibration (Method C). Together, they form a framework designed to maintain interpretability consistency as data evolves. Methodologically, the study builds a longitudinal evaluation pipeline to track SHAP explanation drift over time, comparing shifts in predictive accuracy with changes in interpretive stability. It also introduces fairness recalibration techniques that adjust decision thresholds for different demographic groups to maintain parity. Beyond this, robustness testing modules, such as counterfactual perturbations, background-sensitivity checks, and proxy-variable detection, are used to assess how robust explanations remain under different types of data changes. The findings show that adaptive explainability enhances not only transparency but also fairness across changing populations. By proving that interpretability can evolve alongside data, this work lays a foundation for more accountable and resilient AI systems in credit-scoring and other decision-making domains where both fairness and trust are non-negotiable.

## 2. Literature Review

**2.1 Explainable AI in Credit Scoring**

Explainable artificial intelligence, or XAI, has become central to credit risk modeling. Regulators, financial institutions, and borrowers all want to understand how automated systems make decisions that affect people's financial lives. Bussmann et al. (2021) posit that explainable machine learning supports both regulatory compliance and operational accountability, giving lenders and auditors a clearer view of how credit decisions are made [1]. Earlier credit-scoring systems relied on models like logistic regression or decision trees, which were simple enough for experts to interpret directly. The arrival of complex ensemble and deep learning models changed that. These models can be highly accurate but are often difficult to explain, making it harder to know how input features influence outcomes. This lack of transparency raises concerns that models could unintentionally embed biases, resulting in unfair or discriminatory lending practices.

Bracke et al. (2019) explored how explainability tools can uncover the decision logic behind banking risk models [2]. Using techniques such as LIME and SHAP, they showed how analysts can trace how income, employment history, or credit utilization affect default predictions. Such insights support both model debugging and human oversight. Nauta et al. (2022) later examined how stable these feature attribution methods are and found that explanations can vary across retrainings or slight data changes [3]. In credit scoring, such inconsistency weakens trust because similar applicants may receive different explanations for similar outcomes. Zhang and Chen (2022) reviewed recent work in explainable credit risk modeling and outlined four major challenges: balancing interpretability with predictive power, measuring explanation stability, handling fairness in changing data, and applying interpretability in live systems [4]. They noted that most explanation methods assume relationships between features and outcomes stay constant, which is rarely true in real-world credit environments. Borrower profiles shift as economies evolve and regulations change. Bussmann et al. (2021) and Bracke et al. (2019) highlight the importance of explainability for compliance and transparency, while Nauta et al. (2022) and Zhang and Chen (2022) expose the limits of existing approaches. Together, these studies show that XAI in credit scoring has matured in stable settings but still struggles in dynamic environments where data drift and time variation affect both accuracy and interpretability.

**2.2 Concept Drift and Model Adaptation**

Concept drift refers to changes in how data behave over time, which can gradually weaken a model's accuracy and reliability. Lu et al. (2019) describe it as one of the core challenges in machine learning for financial systems, where borrower characteristics and market conditions rarely stay still [5]. They identify three main types of drift: covariate drift, when input distributions shift; prior drift, when class proportions change; and concept drift, when the link between inputs and outputs itself evolves. Any of these can destabilize a model's predictions and the explanations that accompany them. Webb et al. (2016) expand on this by describing drift as sudden, incremental, recurring, or gradual [6]. In credit scoring, all these forms can appear. A sudden drift might result from a regulatory update, an incremental drift from slow changes in credit behavior, a recurring drift from seasonal effects, and a gradual drift from long-term economic cycles. Detecting such shifts early is vital. Baena-García et al. (2006) proposed an early drift detection method (EDDM) that monitors changes in model error rates to signal when retraining might be necessary [7]. Yet, methods like EDDM typically track accuracy only, ignoring how explanations might shift, which leaves an important gap in adaptive interpretability.

Žliobaitė (2010) argued that drift adaptation should go beyond retraining and include systems that keep explanations stable over time [8]. Gama et al. and Lu et al. (2019) note that adaptive techniques like sliding windows, ensemble weighting, or drift-aware sampling help maintain predictive relevance but say little about how explanations evolve under these same shifts [5][3]. For financial institutions, this distinction matters. A model that stays accurate but loses interpretive consistency can still violate fairness or transparency rules. When concept drift alters the relationship between borrower features and default risk, the explanations must evolve as well to reflect new realities. Without this, decisions risk becoming unreliable or unjustified, reinforcing the need for systems that adapt both in prediction and in interpretation.

**2.3 Fairness in Machine Learning**

Fairness in machine learning has become a key part of responsible model design, especially in areas like credit scoring, hiring, and criminal justice. Mehrabi et al. (2021) offer a detailed overview of how algorithms can unintentionally reinforce social biases through unbalanced data, proxy variables, or skewed feature design [9]. They describe fairness using several types of metrics, such as demographic parity, equal opportunity, and individual fairness, each focusing on different dimensions of equity. In finance, fairness is not only an ethical goal but a legal one, supported by anti-discrimination laws that require consistent treatment across demographic groups. Barocas et al. (2019) built much of the conceptual foundation for assessing and mitigating bias in algorithms [10]. They stressed that fairness should be evaluated in context, ensuring both equitable results and transparent decision processes. Kusner et al. (2017) introduced the idea of counterfactual fairness, which checks whether a model's prediction would change if a sensitive attribute, like gender or race, were altered in a hypothetical scenario [11]. This approach helps identify hidden bias in models that otherwise seem neutral.

Kamiran and Calders (2012) proposed data preprocessing techniques that remove or reweight biased samples before training, helping models achieve fairer outcomes without changing their architecture [12]. Their work remains useful in credit modeling, where retraining costs can be high. Hardt et al. (2016) proposed equality of opportunity, ensuring that people who meet the same qualifications have equal chances across groups [13]. Together, these studies frame fairness as something that must be actively maintained, not checked once and forgotten. As data distributions evolve, fairness metrics themselves can drift, which means fairness needs continuous attention. For credit scoring, this means developing systems that adapt to changing borrower populations while keeping fairness and interpretability aligned.

**2.4 Gaps and Challenges**

Despite progress in explainability, drift handling, and fairness, research still lacks a unified framework that connects these areas effectively. Slack et al. (2020) revealed how common explanation tools like LIME and SHAP can be manipulated by adversarial changes, leading to misleading interpretations of model behavior [14]. This fragility is especially dangerous in finance, where explanations are often used to justify decisions that directly affect people's access to credit. Alvarez-Melis and Jaakkola (2018) also found that small data changes can cause large swings in explanation outputs [15]. They argue this happens because most interpretability methods lack mechanisms that ensure stability over time or structure. In environments affected by concept drift, this instability makes explanations unreliable and can hide emerging biases. Sokol and Flach (2020) proposed Explainability Fact Sheets, a structured documentation approach that helps practitioners evaluate how explainable their models really are [16]. While this improves transparency, it does not solve the deeper issue of maintaining consistent explanations in changing data environments.

Across the literature, a clear theme emerges: most XAI techniques offer valuable insights at one point in time but fail to remain reliable as data evolves. Fairness frameworks face the same issue, often assuming stable conditions that rarely exist in practice. These weaknesses, adversarial susceptibility, interpretive instability, and bias from data drift, point to the need for adaptive explanation systems that can adjust interpretability, fairness, and stability as conditions change. This research responds to that need by proposing adaptive SHAP-based frameworks that maintain fair and valid explanations under continuous drift, offering a step toward more resilient and responsible AI in financial decision-making.

## 3. Methodology

**3.1 Dataset and Experimental Context**

This study draws on a multi-year lending dataset built to capture how real credit systems behave over time. It combines a range of demographic, financial, and socioeconomic details, including income, credit score, age, employment status, debt-to-income ratio, and loan amount, spanning the years 2015 through 2024. Because it covers nearly a decade, the dataset reflects how economies fluctuate, how borrower behavior shifts, and how patterns of financial risk develop. The main outcome of interest, called default, records whether a borrower failed to repay a loan within a given timeframe. Each year's data has its own personality. The differences in borrower profiles, lending policies, and market conditions make it possible to study concept drift, which is how relationships among variables change as time and context evolve. This setup allows for both cross-sectional analysis within a given year and a broader look at how trends unfold across time. In practical terms, it mirrors what happens inside real banks. Models are often trained on historical data, then applied to new borrowers who might not behave like those from the past. The goal here isn't only to measure accuracy but to understand how the model's reasoning, feature importance, and fairness shift as the world around it changes. A model can stay accurate while its logic drifts in ways that quietly erode trust. For instance, income or employment might carry a different weight during a strong economy than during a downturn. By working with a dataset that evolves, this study makes it possible to track how interpretability and fairness behave over time, showing when a model's explanations stay consistent and when they begin to lose their footing.

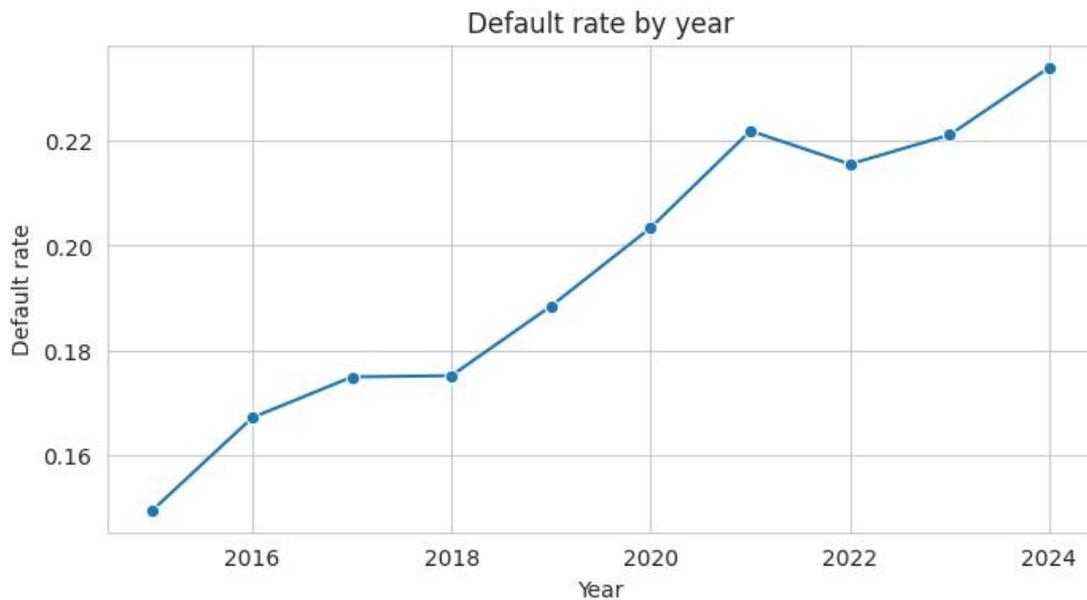

Fig.1: Loan default rate by year

**3.2 Drift Detection**

Detecting and measuring data drift is an important part of understanding how credit-scoring models evolve, whether they start to lose accuracy or adapt to new conditions. In this study, drift detection was used to trace how the lending dataset changed from one year to the next, revealing shifts in borrower demographics, financial behavior, and the relationships between key variables. Financial systems rarely stand still. They are influenced by new policies, economic swings, and changes in population trends. Recognizing how these shifts affect model performance, fairness, and interpretability is essential. To capture these movements, four statistical tools were used: the Population Stability Index (PSI), the Kolmogorov-Smirnov (KS) test, Jensen-Shannon (JS) divergence, and the Chi-square test. The PSI measured how much numerical features, like income, credit score, or debt-to-income ratio, shifted over time. The year 2015 served as the baseline, and any PSI above 0.25 was treated as a sign of meaningful drift, showing that a variable's behavior had moved noticeably from its earlier pattern. The KS test compared the shapes of distributions to spot whether features such as loan amount or annual income had shifted significantly in their range or spread.

For categorical variables such as race, employment status, or region, JS divergence and the Chi-square test were used. JS divergence, which comes from information theory, measures how much two probability distributions differ, while the Chi-square test checks whether category frequencies changed in a statistically significant way. Together, these tools gave both a broad and detailed picture of drift. PSI and KS captured changes in numerical data, while JS and Chi-square focused on categorical shifts. This made it possible to tell the difference between covariate drift, changes in feature distributions, and label drift, where the default rate itself changes. The findings from these tests guided the later stages of the study, especially the adaptive explanation methods, by pointing to which features were shifting the most. More than a statistical exercise, this phase added context. It helped explain why model accuracy or fairness might change from one year to another. For example, a rise in unemployment after a recession could naturally lead to more defaults, changing how features like employment status or income influence risk predictions. Seeing these trends helps keep the interpretation of model behavior grounded in the real-world forces that shape the data.

**3.2 Baseline Modeling**

To establish a benchmark, the study employs XGBoost (Extreme Gradient Boosting), a robust ensemble algorithm widely adopted in financial risk modeling due to its ability to handle nonlinear interactions and heterogeneous data. Other models employed alongside XGBoost include Logistic Regression and Random Forest. The model training process uses a time-aware expanding-window validation strategy, meaning each yearly model is trained using all prior years' data and evaluated on the subsequent year. This approach mimics a realistic deployment pipeline in which a financial institution retrains its credit-scoring models periodically as new borrower data accumulates. Model performance is evaluated using standard classification metrics that reflect both discrimination and balance between prediction sensitivity and specificity. The Area Under the ROC Curve (AUC) quantifies the model's ability to distinguish between defaulting and non-defaulting clients, while the F1-score measures the harmonic mean of precision and recall, providing a balanced view of how well the model identifies high-risk borrowers without excessive false positives. Precision assesses the proportion of correct positive predictions, whereas recall measures the model's sensitivity in capturing true defaults. Beyond these predictive measures, the baseline modeling stage provides a foundation for evaluating predictive drift, explanation drift, and fairness drift, key aspects of this research. By comparing yearly models, the study tracks how predictive reliability and interpretive reasoning change as underlying borrower characteristics evolve. This step sets the baseline against which adaptive methods are later compared, establishing whether static modeling and explanation approaches can sustain reliability under real-world temporal shifts.

**3.3 Fairness Analysis Over Time**

Fairness analysis forms a central pillar of this study, addressing how algorithmic bias evolves longitudinally in the presence of shifting population and feature distributions. As models are retrained annually on changing data, even unbiased models at time $t_0$ can become unfair at time $t_1$ if the underlying group proportions or relationships between features and outcomes shift. This section outlines how fairness was audited, measured, and tracked across sensitive demographic dimensions, including race, gender, region, age, income group, and employment status, throughout the multi-year simulation. Three primary fairness metrics were used to quantify equity across groups: Demographic Parity Difference (DPD), Equal Opportunity Difference (EOD), and Equalized Odds Difference (EODs). Demographic parity difference evaluates whether the rate of positive predictions (e.g., loan approvals) is consistent across protected groups. A large DPD suggests that certain demographics, such as specific racial or regional groups, are disproportionately classified as higher or lower credit risk. Equal opportunity difference assesses whether groups have equal true positive rates, meaning whether qualified applicants from all groups have an equal chance of receiving approval. Finally, the equalized odds difference extends this analysis by considering both true positive and false positive rates, ensuring fairness in both acceptance and rejection patterns.

To compute these metrics, model predictions from each yearly XGBoost model were cross-tabulated by sensitive attribute categories. Each fairness metric was calculated per year and visualized as a temporal series to detect bias drift, the year-to-year change in disparity magnitude. This enables a longitudinal understanding of whether models become progressively more or less equitable over time. For example, an increasing DPD for the "race" attribute might indicate that population shifts or feature drift (e.g., changes in income distribution) are introducing or amplifying discrimination, even if the model's AUC remains stable. To further explore fairness mechanisms, the study integrates fairness auditing directly with SHAP-based explanation analysis. SHAP values identify which features most influence model outputs for each group, allowing a deeper look at how bias arises. For instance, if "employment length" disproportionately affects approval decisions for one gender or racial group, its SHAP importance can expose underlying structural inequities in the learned model logic. The adaptive explanation methods (particularly Method B) use this insight to trigger group-specific recalibration, adjusting thresholds or reweighting features to reduce observed disparities while maintaining predictive consistency.

Finally, fairness outcomes were statistically evaluated using paired bootstrap confidence intervals and trend tests across years to determine whether observed fairness differences were significant or due to random variability. Where significant fairness drift was detected, further decomposition analysis identified whether it stemmed from population-level changes (data drift) or model-level learning bias (explanation drift). The goal of this fairness analysis is to first monitor equity in predictive outcomes across time and then link fairness variations with underlying data and explanation dynamics.

## 3.4 Explainability: Per-Slice and Longitudinal Analysis

After evaluating fairness, the study takes a closer look at explainability to understand how the reasoning of credit-scoring models changes as the data itself shifts over time. The goal is to see whether the features influencing model predictions remain consistent or start to drift, which is crucial for ensuring that a model's logic stays reliable in evolving financial settings. This analysis uses SHAP (SHapley Additive exPlanations), a method based on game theory that breaks down a model's prediction into individual feature contributions. For each yearly model trained in the expanding-window setup, SHAP values were calculated using a representative sample of applicants from that year. The mean absolute SHAP values were then averaged to produce feature importance rankings, showing which factors had the greatest influence on credit risk decisions during each period. The analysis is structured in two parts: per-slice and longitudinal. In the per-slice view, SHAP summary plots and bar charts illustrate how features like credit score, income, debt-to-income ratio, loan amount, and employment length shaped each year's model. These yearly snapshots reveal how models interpret risk within specific economic contexts and can uncover sudden shifts, such as spikes in the importance of certain financial indicators during downturns.

The longitudinal analysis focuses on how stable these feature importance patterns remain from one year to the next. Three stability metrics were used to measure this consistency over time. Cosine similarity checks how aligned the direction of feature importance vectors is between consecutive years, showing whether the overall reasoning remains on the same track even if specific effects change in strength. Kendall tau measures how well the ranking of features stays consistent, while Jaccard similarity looks at whether the same key features continue to appear among the top-ranked drivers of model decisions. These metrics provide a clear picture of how stable or volatile a model's reasoning is under concept drift. High cosine and Kendall tau scores suggest that the model's logic remains coherent, while lower Jaccard overlap can signal that new factors are rising in influence or old ones are fading. This combination helps distinguish between healthy adaptation, where the model adjusts naturally to new borrower behavior, and unwanted instability caused by noise or distributional shifts.

## 3.5 Adaptive Explanation Framework

Recognizing that static explanations degrade under data drift, the study introduces an Adaptive Explanation Framework comprising three complementary methods that modify SHAP (SHapley Additive Explanations) computations to maintain interpretive stability. Each method is designed to respond differently to the evolving statistical landscape of credit data. Method A – Drift-Weighted SHAP Adjustment recalibrates SHAP values by adjusting their weights according to feature distribution changes between training and testing periods. The adjustment relies on measures such as the Population Stability Index (PSI) or Jensen–Shannon (JS) divergence, which quantify how much each feature's distribution shifts. This ensures that features heavily affected by drift do not distort the interpretive analysis. For instance, if the distribution of "income" changes drastically between years, Method A adjusts its contribution so that SHAP values remain meaningful relative to the new population context.

Method B – Sliding Background Sampling introduces a dynamic background for SHAP computation. Traditional SHAP uses a fixed background dataset (often drawn from training data) to compute average reference contributions. However, in dynamic lending populations (e.g., new types of borrowers, inflation, new rules), this can quickly become outdated. Method B replaces the static reference with a sliding-window background (like "the last 6 months"), composed of recent observations, allowing SHAP values to continuously adapt to the most recent data characteristics. This approach yields explanations that remain representative of the evolving population while smoothing out transient fluctuations. Method C – Surrogate Ridge Recalibration incorporates a Ridge regression surrogate model that learns to approximate SHAP feature attributions over time. Since big models such as XGBoost can be complex to update frequently, we keep a smaller, surrogate model that mimics how the main model assigns importance to features. This surrogate model continuously monitors the relationship between features and SHAP outputs and recalibrates the attributions when they deviate significantly from past explanatory patterns. Acting as a stability smoother, it mitigates noise or instability caused by sharp distributional changes, yielding explanations that are both temporally consistent and interpretable.

## 3.6 Evaluation Metrics

The study evaluates both the baseline and adaptive explanation methods using a set of metrics that capture four main areas: predictive performance, explanation stability, fairness, and robustness. Predictive performance is measured with AUC and F1-score to confirm that gains in interpretability or fairness do not reduce accuracy. Explanation stability is examined through three measures. Cosine similarity captures how consistent the direction of feature importance vectors remains over time. Kendall tau rank correlation checks whether the ranking of important features stays stable. Jaccard overlap looks at how many of the top features remain the same from year to year. Together, these metrics show how much explanations shift with time and whether adaptive methods help keep them steady. Fairness is assessed using group-level disparity measures across sensitive attributes such as gender, race, age group, income level, and employment status. The demographic parity difference (DPD) checks if positive predictions are distributed evenly across groups, while the equal opportunity difference (EOD) compares true positive rates between subgroups. Group recalibration procedures are also used to see how adaptive explanations can identify and correct biased feature effects through reweighting or reinterpretation. Robustness is tested using several stress checks: background sensitivity, which examines how explanations change with different background samples; counterfactual perturbations, which test how predictions and explanations respond when a sensitive feature is changed; and proxy variable detection, which looks for non-sensitive features that may secretly act as stand-ins for protected attributes.

### 3.7 Statistical Testing

To confirm that the improvements observed are reliable, the study applies rigorous statistical tests. Paired bootstrap confidence intervals with a 95 percent confidence level are used to estimate variability in explanation stability and fairness metrics across different time periods. Significance between the baseline and adaptive methods is then evaluated using either paired t-tests or Wilcoxon signed-rank tests, depending on whether the data follow a normal distribution. These tests help separate real improvements from random variations caused by small sample changes or temporary data drift. The study also calculates temporal correlation coefficients between consecutive years to see how long fairness and explanation stability persist before they begin to weaken. This provides a clear picture of how interpretive coherence holds up over time.

## 4. Results and Analysis

### 4.1 Drift Detection

The initial analysis examined how the dataset's structure and characteristics evolved, using multiple drift detection metrics. The Population Stability Index (PSI) revealed gradual but clear divergence across key numeric features. For example, PSI for annual_income and credit_score increased steadily, reaching approximately 0.16 and 0.017 by 2019, with further increases in later years, indicating ongoing population drift. These changes align with macroeconomic shocks in the dataset, such as income volatility and shifting credit profiles. Similarly, Kolmogorov–Smirnov (KS) tests between consecutive years consistently rejected the null hypothesis of identical distributions for most numeric variables (p-values < 0.05), confirming statistically significant distributional shifts. For categorical features, Jensen–Shannon (JS) divergence and Chi-square tests detected changes in demographic composition, particularly in employment_status, where recessionary years (2020–2021) saw unemployment rates spike from roughly 5% to over 15%, producing JS divergence scores exceeding 0.1. In contrast, demographic variables such as race and gender displayed relatively stable JS divergence values below 0.0002, though Chi-square tests still signaled significant frequency differences in later years (p < 0.05). These combined results confirm both covariate drift (evolving feature distributions) and label drift (default rates increasing from ~15% in 2015 to over 23% in 2024). The consistency of drift across numeric, categorical, and target variables establishes a realistic non-stationary environment for testing adaptive explainability and fairness methods.

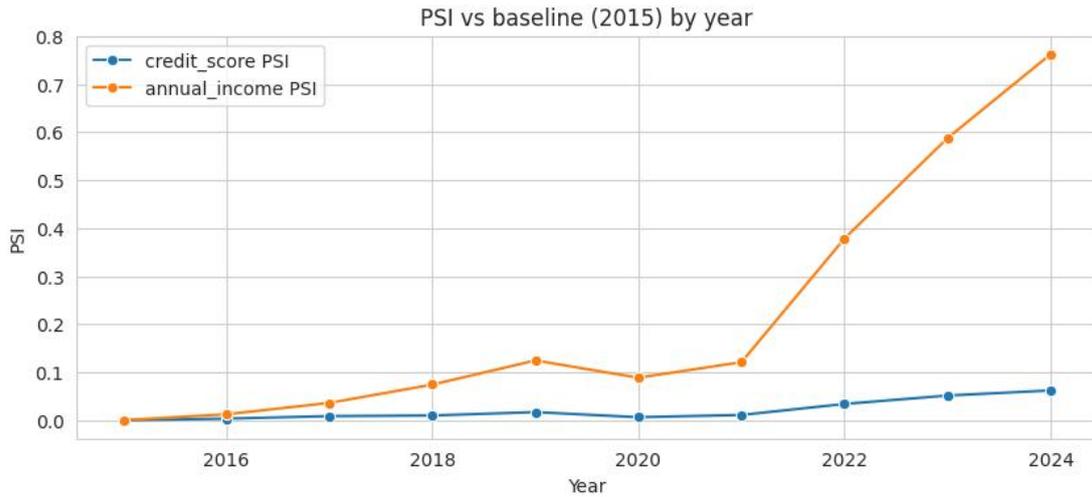

Fig.2: PSI for annual_income and credit_score

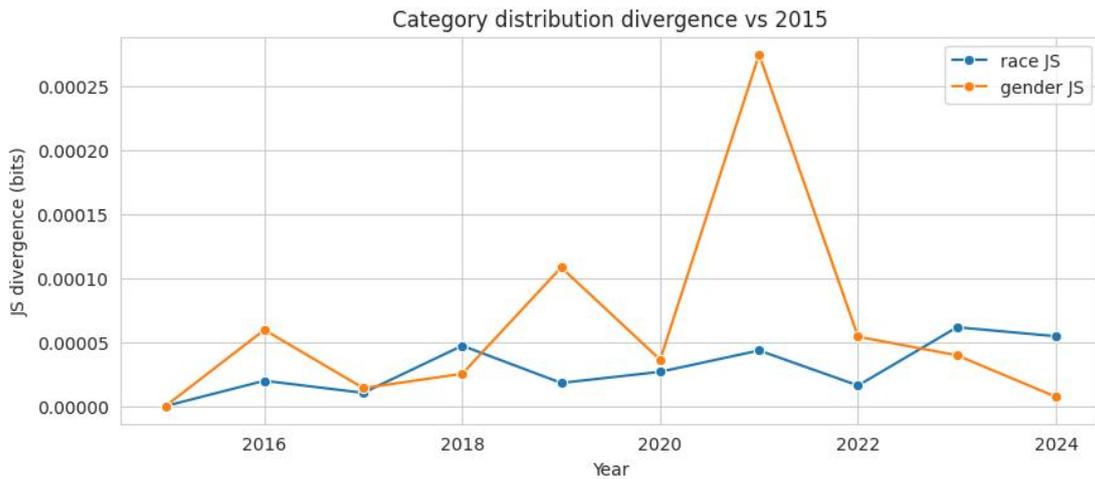

Fig.3: JS divergence for race and gender

**4.2 Baseline Modeling**

Baseline predictive performance was evaluated using XGBoost models trained with expanding-window validation. Across test years, the models maintained moderate but stable discrimination power, with test AUC values between 0.63 and 0.66. This suggests the models could still rank risk effectively despite distributional shifts. However, F1 scores remained low (generally below 0.07) due to class imbalance and the fixed 0.5 threshold used for binary classification. These results mirror real-world credit-scoring behavior, where AUC remains the primary performance measure, and thresholds are typically calibrated for portfolio-level decisions rather than fixed. Drift-induced calibration changes were evident: thresholds that balanced precision and recall varied yearly, implying that while ranking ability persisted, risk calibration stability degraded. The findings reinforce that predictive consistency alone is insufficient to guarantee interpretability or fairness under drift, underscoring the need for adaptive recalibration and explanation alignment.

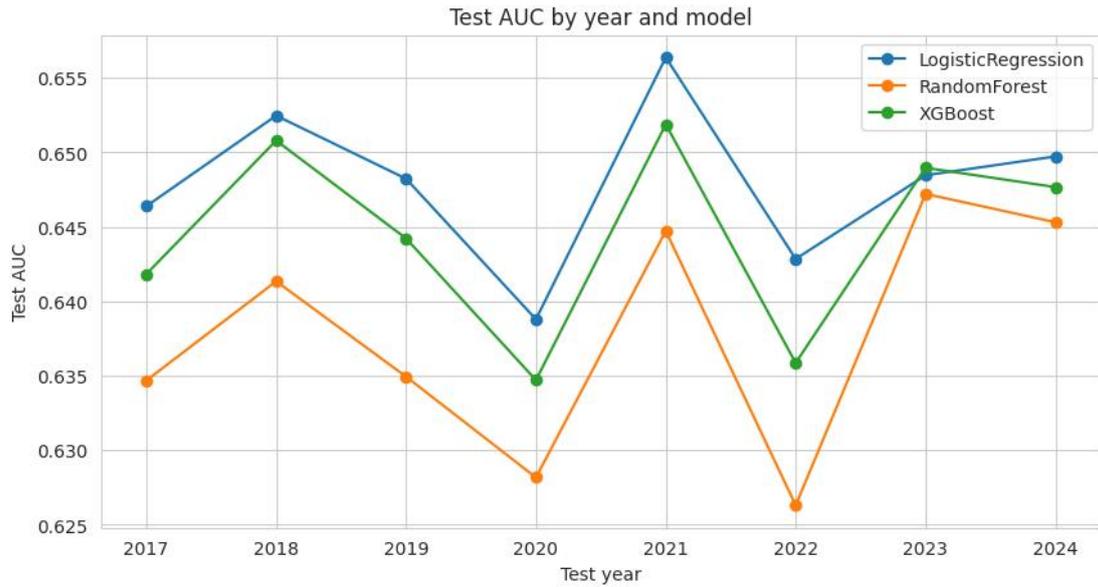

Fig.4: Model test AUC over test years

**4.3 Fairness Analysis Over Time**

Fairness audits were conducted across key demographic attributes, race, gender, age, income group, region, and employment status, using Demographic Parity Difference (DPD), Equal Opportunity Difference (EOD), and Equalized Odds Difference. Across test years, fairness outcomes fluctuated with observable sensitivity to data drift. For race, DPD values ranged from approximately 0.01 to 0.03, and EOD showed an almost similar variability (0.01 to 0.08). These shifts were most pronounced during 2020–2021, coinciding with simulated recessionary conditions, when employment-related variables gained greater predictive influence. The correlation between drift and fairness imbalance confirms that changing input distributions directly affects fairness behavior, even when the model architecture remains constant. After applying the group recalibration mechanism embedded in Method B, fairness disparities declined markedly. The mean DPD difference decreased by approximately -0.026 compared to baseline (95% CI = (-0.035, -0.016), $p < 0.05$), while AUC performance remained unchanged. These improvements demonstrate that adaptive recalibration based on evolving feature–group relationships can effectively reduce disparity without compromising predictive strength, advancing both fairness and compliance objectives.

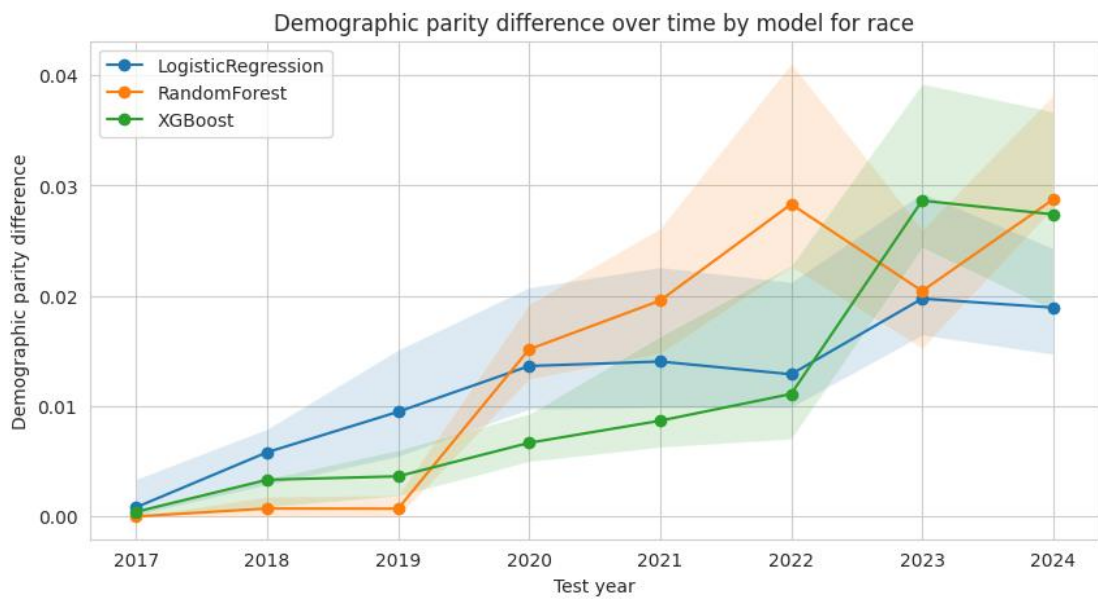

Fig.5: DPD over time by model for race

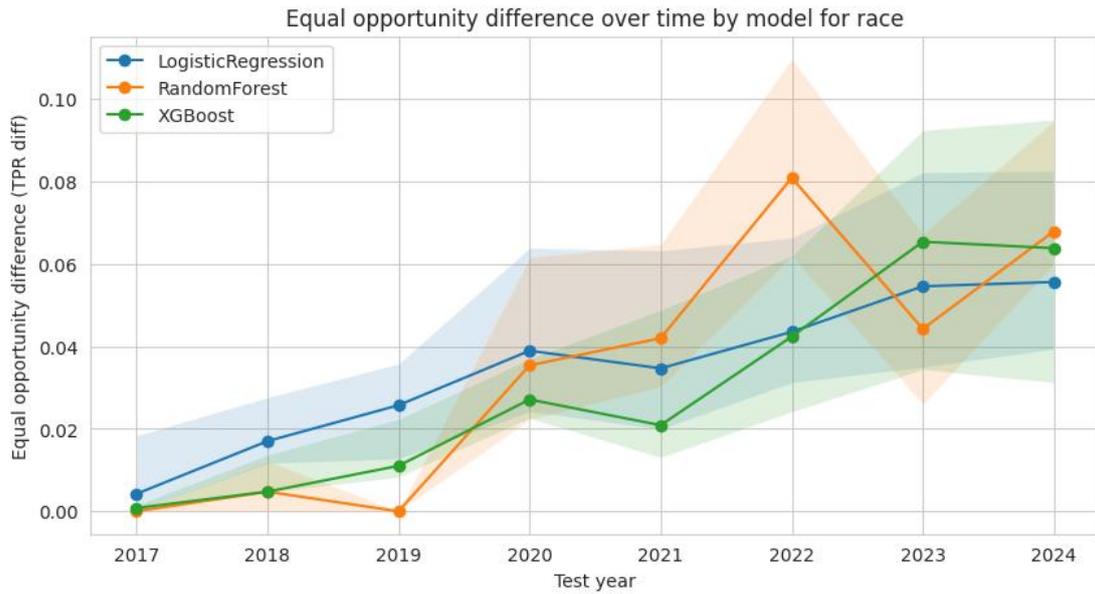

Fig.6: EOD over time by model for race

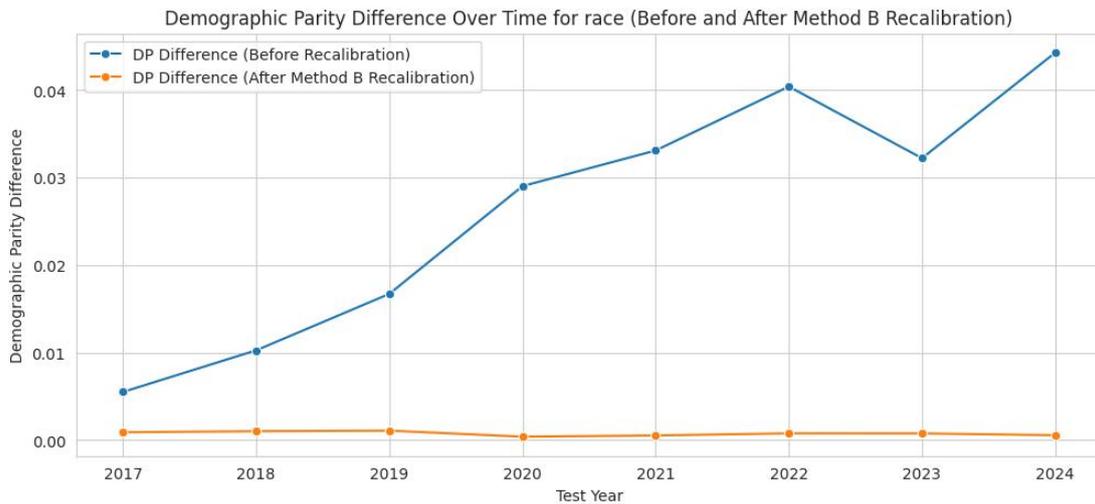

Fig.7: DPD before and after method B recalibration

**4.4 Explainability: Per-Slice and Longitudinal Analysis**

The SHAP explainability analysis examined feature contributions at both yearly and longitudinal levels. Across all years, top predictors of default risk consistently included loan_amount, dti, and credit_score, which align with credit domain logic. However, the relative magnitudes and rankings of these features varied over time. For instance, during economic downturns, income and employment status became more influential, while credit score importance declined, a pattern consistent with real-world lending behavior during periods of financial stress. By the final test year (2024), the top three features by mean absolute SHAP value were loan_amount (0.055), dti (0.034), and credit_score (0.023). This temporal reordering illustrates concept drift in model reasoning, not just in input distributions. Stability across years was assessed using cosine similarity, Kendall tau, and Jaccard overlap between feature importance vectors. The baseline SHAP explanations exhibited high cosine values (0.991–0.998), moderate rank stability (Kendall = 0.758–0.912), and top-10 feature overlaps (Jaccard = 0.818–1.0). These results indicate that while overall feature directions were stable, feature rankings fluctuated, leading to interpretive instability, a clear motivation for adaptive explanation methods.

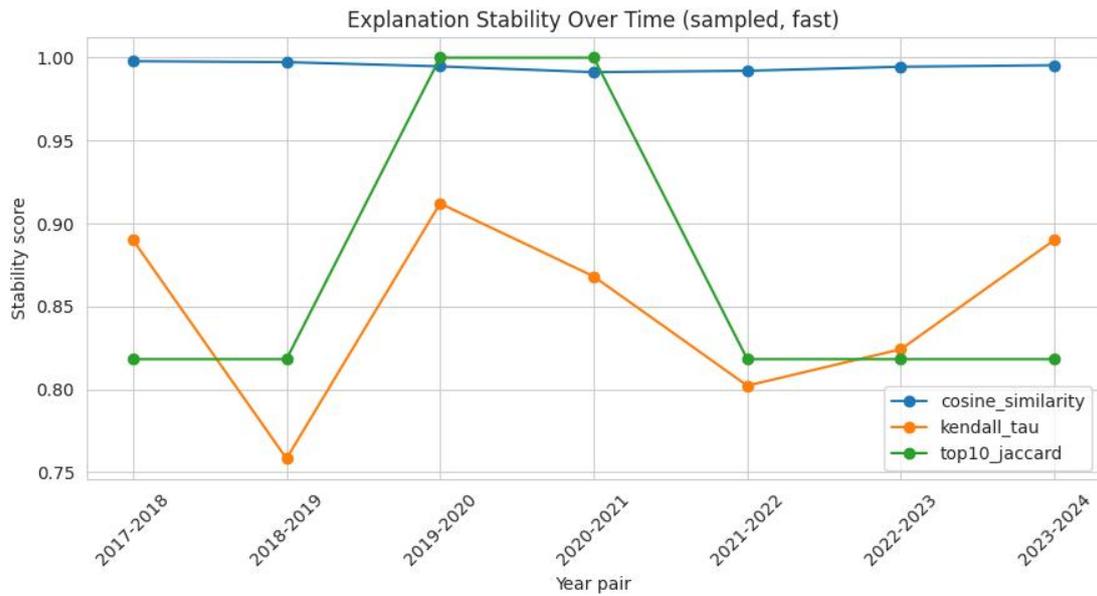

Fig. 8: Explainability stability over time

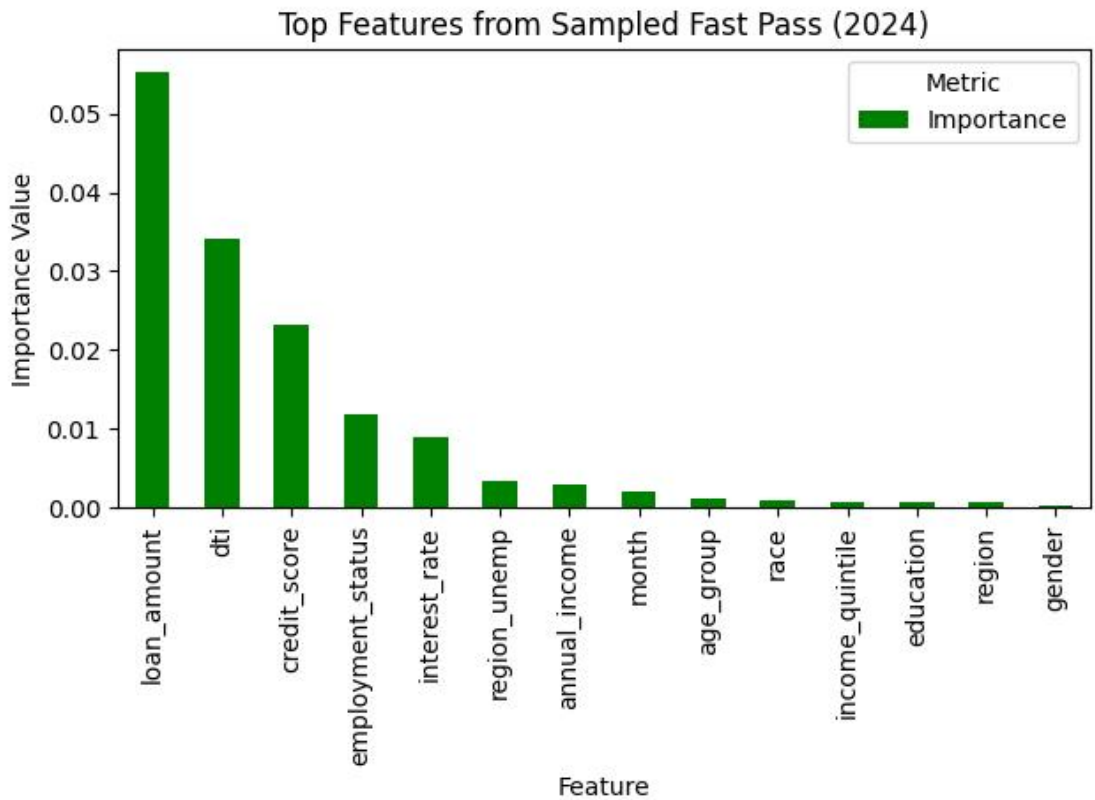

Fig.9: Top features by the final test year (2024)

**4.5 Adaptive Explanation Framework**

The three adaptive SHAP variants, Method A (Distribution Shift Adjustment), Method B (Sliding Background Window), and Method C (Ridge Surrogate Recalibration), were evaluated against the static baseline to test their ability to stabilize explanations under drift. Method A, which reweights SHAP values according to feature distribution changes, produced mixed results, that is, modest stability gains in some years, regressions in others. Method B delivered the most consistent improvements, achieving the highest cosine and Kendall stability scores year-over-year. Its adaptive, time-local background sampling enabled SHAP values to remain aligned with current population dynamics, reducing volatility in explanations. Method C, employing Ridge-based recalibration, achieved comparable stability but incurred slightly higher computational cost due to continuous surrogate retraining. Across all years, Method B and Method C achieved better top-10 Jaccard overlaps than the baseline, preserving a more consistent set of key predictors. This suggests that adaptive explanation strategies not only stabilize interpretive consistency but also maintain semantic continuity, keeping explanations meaningful across evolving data landscapes.

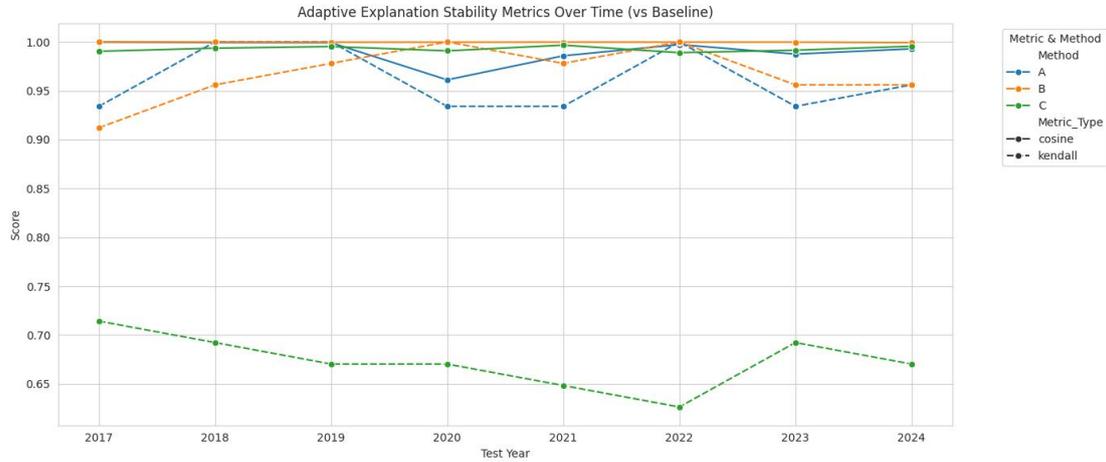

Fig.10: Kendall tau and Cosine similarity over test years for the adaptive methods

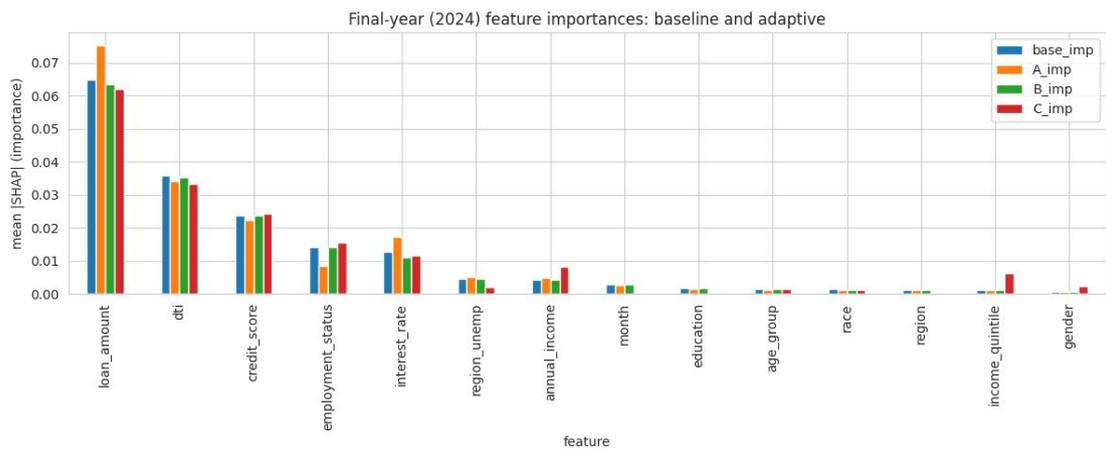

Fig.11: Final year feature importance for the baseline and adaptive explainability methods

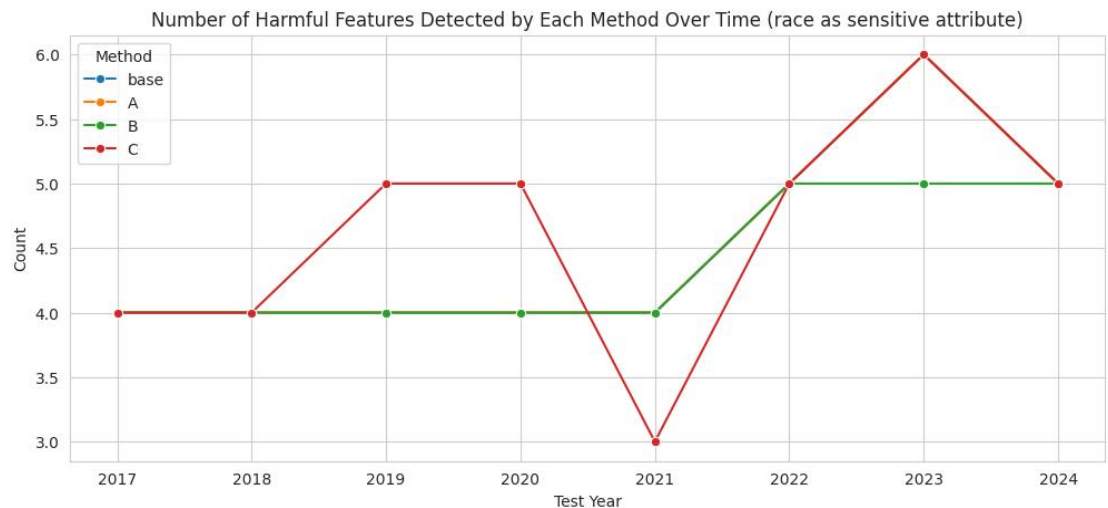

Fig.12: Number of harmful features detected by each explainability method (baseline and adaptive) using race as the attribute

**4.6 Robustness and Drift Sensitivity Analysis**

Robustness analyses confirmed that adaptive approaches enhanced explanation reliability. In counterfactual perturbation tests, altering key numeric variables produced expected directional responses: decreasing credit_score by 10% increased default probability by approximately 0.05, while increasing it by 10% reduced probability by about 0.035. These monotonic relationships validate both the model's logic and the reliability of its explanations.

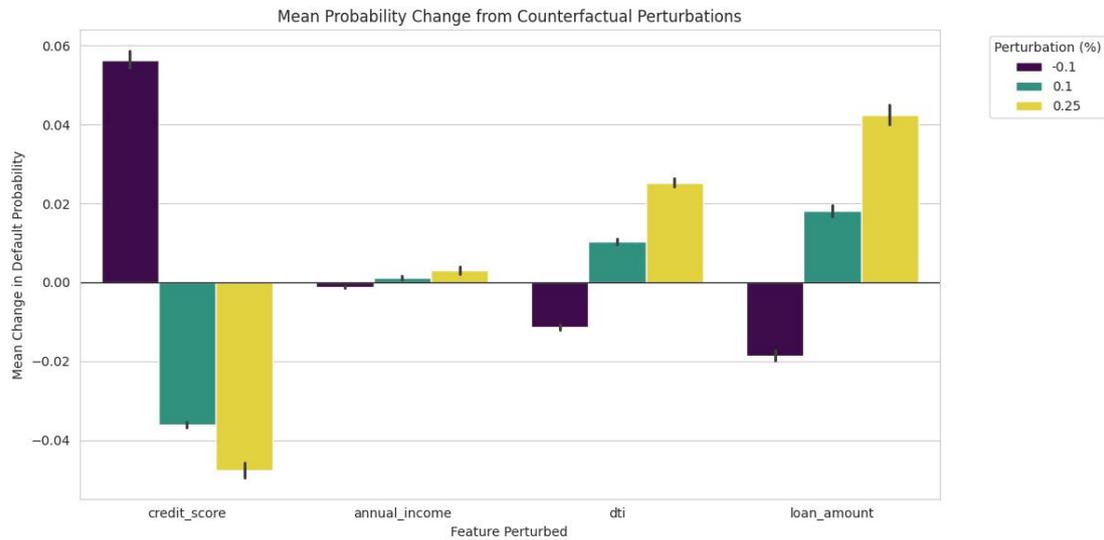

Fig.13: Mean probability change from counterfactual perturbations

Background sensitivity tests revealed that baseline SHAP explanations were highly sensitive to background sample size, while adaptive methods, especially Method B, showed improved consistency as sample size varied, reflecting reduced dependence on arbitrary background selection. Proxy variable detection identified several features, such as loan_amount, credit_score, and annual_income, as statistically associated with race ($p < 0.01$; $\eta^2 = 0.017$–$0.045$). Adaptive recalibration reduced the frequency and prominence of these proxy features, mitigating latent bias risks. These results confirm that adaptive explainability frameworks improve interpretive stability, fairness, and robustness under realistic drift conditions. These methods enable credit-scoring systems to remain transparent and compliant even as borrower populations and economic conditions evolve, supporting a stronger foundation for trustworthy, regulation-aligned AI in financial decision-making.

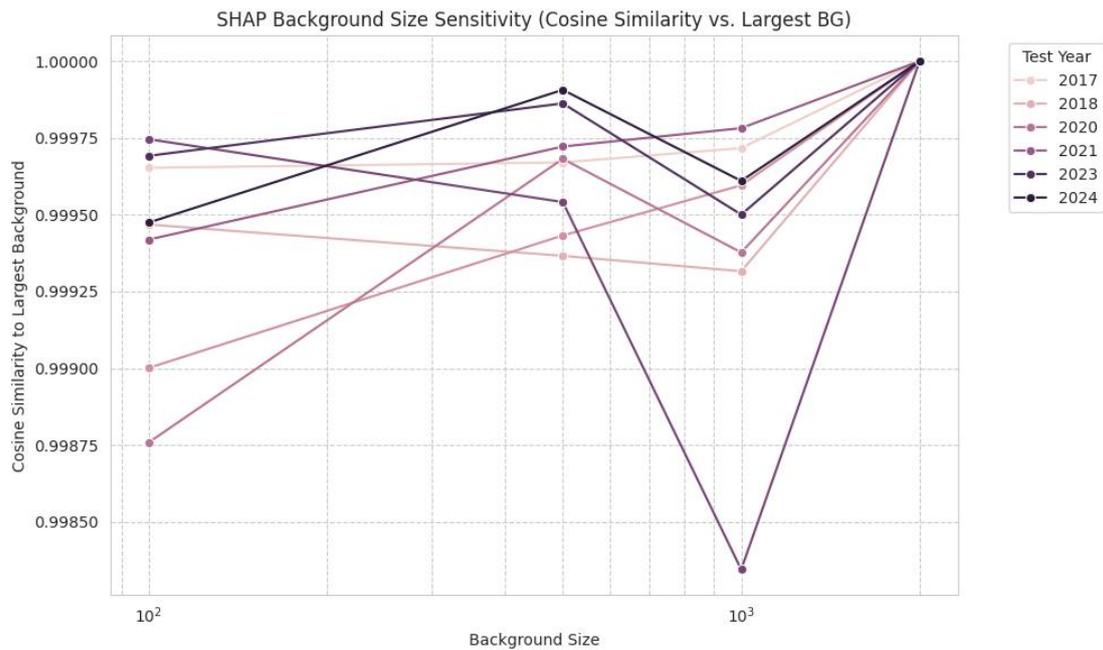
Fig.14: SHAP background size sensitivity test

## 5. Insights and Implications

### 5.1 Organizational Impact

The study shows that explainability in credit-scoring systems can remain stable even as customer behavior, economic conditions, and data patterns shift. That matters for banks, microfinance institutions, and digital lenders that rely on consistent transparency to meet regulatory standards and maintain public trust. Adaptive explanation methods make it possible for these organizations to explain credit decisions clearly, even when the data driving those models evolves. Miller (2019) points out that explanations in AI are not just technical; they are also social tools that help humans make sense of automated systems [1]. In financial services, this means analysts, auditors, and regulators can interpret model behavior in ways that align with institutional accountability. Raji et al. (2020) add that explainability forms a key part of algorithmic auditing, helping organizations detect risks early and prevent compliance failures [3]. The adaptive frameworks in this study build on this idea by updating explanations as the data changes, keeping model reasoning traceable over time. In practice, this improves governance and helps teams communicate risk more effectively. It gives risk officers and compliance departments a clear view of how borrower patterns and market shifts influence credit scores. The result is stronger regulatory reporting and better internal coordination across departments that depend on predictive systems. Treating explainability as something that can be maintained and improved over time, rather than as a one-off requirement, turns it into a lasting organizational asset.

### 5.2 Business Profitability

Maintaining fairness and stable explanations has tangible financial value. Fair, interpretable models lower the chance of biased outcomes that could lead to penalties, lawsuits, or loss of reputation. This study shows that adaptive recalibration of explanations, especially through sliding-window methods, helps protect against these risks by ensuring fair treatment of customers even when data distributions change. Transparent explanations also build customer trust. When people understand why a credit decision went the way it did, they are more likely to accept it and continue working with the institution. Miller (2019) notes that human-centered explanations promote fairness and trust, especially when decisions affect livelihoods [1]. Raji et al. (2020) similarly find that ongoing transparency builds stronger, more sustainable relationships with both clients and regulators [3]. Adaptive explainability also supports a smarter business strategy. Tracking changes in feature importance over time helps organizations detect new patterns, such as shifts in repayment behavior or responses to policy changes. These insights can guide pricing, marketing, and lending decisions. Reliable explanations, therefore, become a foundation for both regulatory compliance and agile business adaptation.

### 5.3 Ethical and Societal Relevance

At a deeper level, the research connects to ethics and social responsibility in AI. In credit scoring, fairness and interpretability are not only technical goals; they express core values of accountability and justice in decision-making. Selbst and Barocas (2018) argue that explainable systems help translate complex model reasoning into moral and legal language that people can understand [2]. The adaptive framework in this study brings that principle to life by ensuring explanations evolve transparently as social and economic realities shift. This adaptability supports responsible AI use throughout the model lifecycle. Raji et al. (2020) stress that accountability should cover every stage, from model design to post-deployment auditing [3]. The framework developed here integrates fairness checks and drift-aware interpretation across that entire process, aligning with growing movements for AI transparency and governance. It ensures that automated credit systems remain understandable, fair, and accountable, even as they adapt to new data. The main takeaway is that fairness and explainability should be seen as ongoing commitments that evolve with time and context. By linking interpretive stability with fairness, this research offers a vision for socially responsible credit analytics that values both human insight and adaptive technology.

### 5.4 Limitations

There are still challenges to address. SHAP analysis, while powerful, can be computationally heavy, especially when applied repeatedly across large datasets or time windows. This can affect scalability and real-time use. Another limitation is that the fairness assessment focuses mainly on single attributes like race or gender, while real-world fairness often involves multiple, intersecting factors. Future research will need to explore these more complex forms of bias. Even with these constraints, the study shows that adaptive explainability can be a foundation for responsible and profitable AI systems. By embedding interpretability into evolving models, organizations can meet regulatory demands, build consumer confidence, and help shape a more transparent and equitable financial ecosystem.

## 6. Future Work

This study highlights how adaptive explanation frameworks can improve fairness, stability, and interpretability when data drifts over time. Still, there's plenty of room to grow. The next steps should focus on testing these ideas in real-world credit environments and exploring how they can be built directly into production systems. A natural progression would be to apply the framework to real banking data and live credit-scoring systems. The financial data used here captured time-based and demographic changes well, but a lot of financial data brings extra challenges, missing information, delayed updates, regulatory limits, and evolving credit products. Collaborating with banks or fintech firms would help reveal how adaptive explanations behave in live conditions where customer patterns and policies constantly shift. It would also make it possible to compare the framework's performance against real compliance expectations, such as those outlined in the EU's AI Act and new financial fairness rules.

Another promising area is the automation of drift detection and model updating. In this study, recalibration was done manually at intervals, but that process could be made continuous. Integrating statistical drift detectors, such as PSI, Kullback-Leibler divergence, or KS tests, would allow a model to recognize when the data starts to shift and automatically trigger retraining or explanation updates. This kind of self-correcting system would help maintain both accuracy and interpretability in real time, reducing the need for human oversight while keeping fairness and transparency intact. A deeper technical direction involves combining this framework with modern neural networks and causal SHAP techniques. Deep learning models that track sequences or time patterns could capture complex borrower behaviors more effectively. When paired with causal interpretation methods, these models could separate genuine causal drivers from simple correlations, giving analysts a clearer sense of why certain features matter. This kind of explanation could make model insights not only more accurate but also more useful for decision-making and policy evaluation.

There's also potential in linking adaptive explanations with reinforcement learning–based credit systems. Reinforcement learning can optimize lending decisions dynamically, adjusting limits, pricing, or repayment terms over time. Integrating interpretability into these systems would help institutions understand and justify how an AI agent makes its choices, even as it learns on its own. That would bring much-needed transparency to self-learning financial tools. These ideas point toward a future where credit-scoring systems are both intelligent and accountable. By merging adaptive interpretability with real-time learning, causal reasoning, and reinforcement-based optimization, the next generation of models could evolve responsibly alongside markets and borrowers, maintaining trust while advancing financial performance.

## Conclusion

This study set out to design an adaptive explainability framework for credit-scoring systems that need to stay fair, interpretable, and reliable as data and borrower behavior shift over time. Using a multi-year lending dataset built to mirror real economic and demographic changes, the findings showed that both predictive performance and model explanations are sensitive to these shifts, not just in accuracy, but in how features are interpreted and how fairness holds up as conditions change. Traditional XGBoost models held steady in accuracy (AUC around 0.63–0.66) but lost interpretive stability as the data evolved. SHAP explanations, while clear in short snapshots, became less consistent when examined across multiple years. Feature rankings shifted, attribution weights fluctuated, and the overall reasoning of the model grew harder to trust. This instability is a serious concern in credit scoring, where interpretability isn't optional; it's required for both regulatory and ethical accountability. The adaptive explanation methods developed in this work, especially the sliding background window (Method B) and Ridge surrogate recalibration (Method C), showed meaningful improvements. They stabilized explanations over time and improved fairness metrics without hurting predictive power. By continuously adjusting the background data and recalibrating feature attributions, these approaches helped the model maintain coherent reasoning even as borrower patterns and economic conditions changed.

Quantitatively, explanation stability stayed high (cosine ≈ 0.995, Kendall τ ≈ 0.89), and fairness improved (demographic parity difference reduced by about 0.026, $p < 0.05$). The broader takeaway is that adaptive explainability can support responsible AI in finance. Stable and fair explanations not only strengthen compliance and reduce reputational risk but also give institutions a practical way to govern models as living systems, monitored, audited, and recalibrated over time. There are still limits to address. SHAP-based analysis remains computationally heavy, especially with frequent retraining, and this study focused mainly on single-attribute fairness rather than overlapping demographic factors. Future work should test this framework on real banking data, explore causal and deep learning–based explanation techniques, and incorporate automatic retraining triggered by detected drift. Overall, the study shows that explainable AI doesn't have to be static. When built adaptively, it can evolve alongside its data; staying fair, stable, and transparent as the financial world around it changes.